\documentclass{article}

\PassOptionsToPackage{numbers}{natbib}

\usepackage[preprint]{neurips_2022}

\usepackage[utf8]{inputenc} 
\usepackage[T1]{fontenc}    
\usepackage{hyperref}       
\usepackage{url}            
\usepackage{booktabs}       
\usepackage{amsfonts}       
\usepackage{nicefrac}       
\usepackage{microtype}      
\usepackage{xcolor}         

\usepackage{amsmath}
\usepackage{amssymb}
\usepackage{amsthm}
\usepackage{cleveref}
\usepackage{graphicx}
\usepackage{caption}
\usepackage{subcaption}
\usepackage{simpler-wick}

\newcommand{\id}{\texttt{id}}

\title{Grokking modular arithmetic}

\author{%
    Andrey Gromov 
    \\
    Meta AI \\
    Meta Platforms, Inc. \\
    Menlo Park, California 94025 \\
    \ \& \\
    Department of Physics,\\
    Condensed Matter Theory Center,\\
    University of Maryland\\
    College Park, Maryland  20740 \\
    \texttt{gromovand@meta.com}\\
}

\begin{document}

\maketitle

\begin{abstract}
    We present a simple neural network that can learn modular arithmetic tasks and exhibits a sudden jump in generalization known as ``grokking''. Concretely, we present (i) fully-connected two-layer networks  that exhibit grokking on various modular arithmetic tasks under vanilla gradient descent with the MSE loss function in the absence of any regularization; (ii) evidence that grokking modular arithmetic corresponds to learning specific feature maps whose structure is determined by the task; (iii) analytic expressions for the weights -- and thus for the feature maps -- that solve a large class of modular arithmetic tasks; and (iv) evidence that these feature maps are also found by vanilla gradient descent as well as AdamW, thereby establishing complete interpretability of the representations learnt by the network.
\end{abstract}

\section{Introduction and overview of literature}

\begin{figure}[!h]
    \centering
    \includegraphics[width=\textwidth]{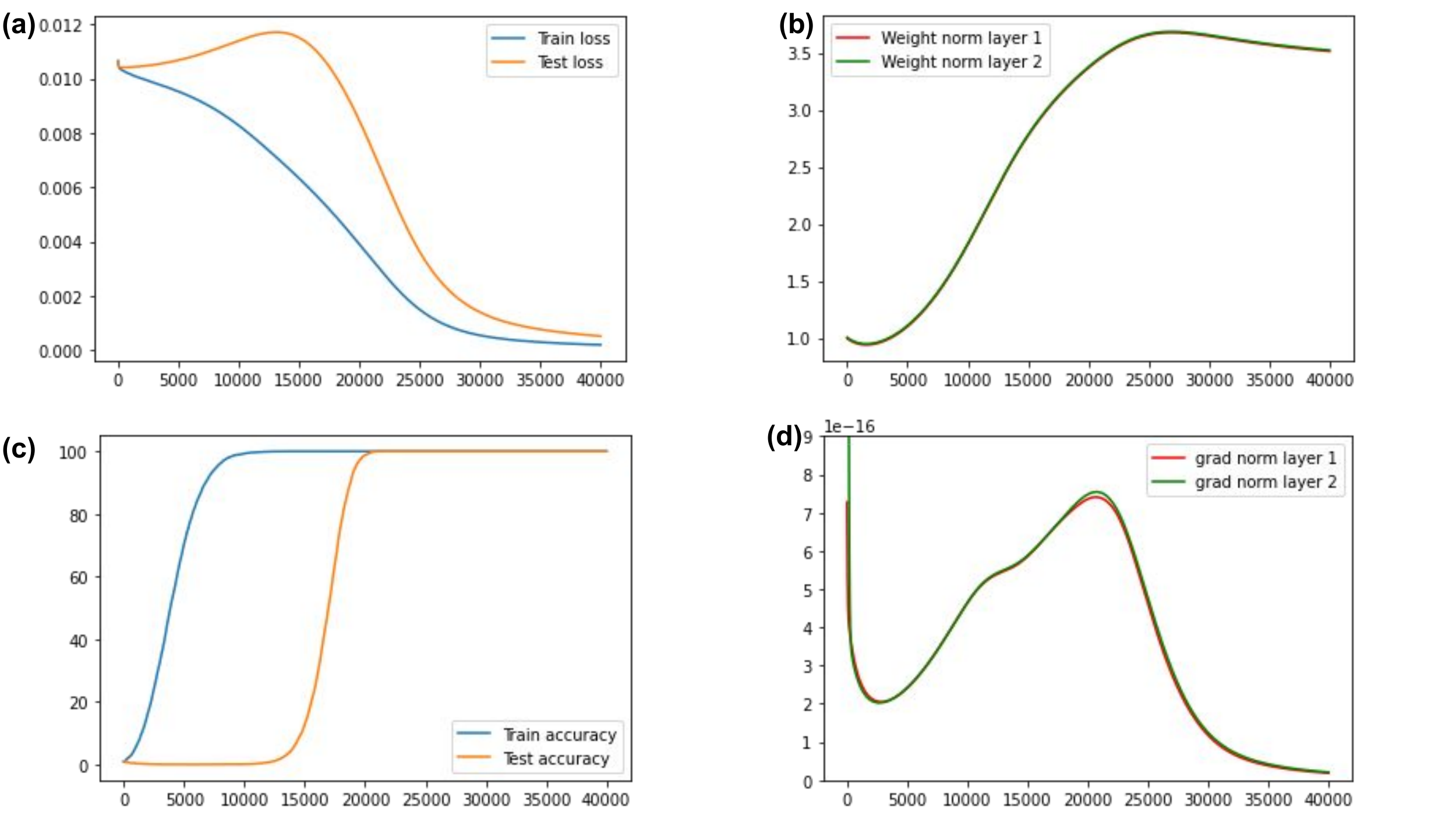}
    \caption{Dynamics under GD for the minimal model \eqref{eq_f} with MSE loss and $\alpha = 0.49$.
    \textbf{(a)} Train and test loss. Train loss generally decays monotonically, while test loss reaches its maximum right before the onset of grokking.
     \textbf{(b)} Norms of weight matrices during training. We do not observer a large increase in weight norms as in \cite{thilak2022slingshot}, but we do see that weight norms start growing at the onset of grokking. \textbf{(c)} Train and test accuracy showing the delayed and sudden onset of generalization. \textbf{(d)} Norms of gradient vectors. The dynamics accelerates until the test loss maximum is reached and then slowly decelerates.}
    \label{fig:fig0}
\end{figure}

Grokking is an effect discovered empirically in \cite{power2022grokking}. Its phenomenology is characterized by a steep and delayed rise in generalization from $0\%$ to a fixed value, as depicted in Fig.~\ref{fig:fig0}b. Beyond that observation, however, there are no clear characteristics of grokking that are reproduced across different works. Here, we start with a lightening review of various claims made in the literature.

In the original work \cite{power2022grokking}, the authors studied how a shallow transformer learns data distributions that are generated by simple deterministic rules  (termed `algorithmic datasets'). Examples of such datasets include modular arithmetic, finite groups, bit operations and more. Specifically, in \cite{power2022grokking} the data took a form of a string ``$a \circ b = c$'', where $c$ was masked and had to be predicted by a two-layer decoder-only transformer. In that study, the following empirical facts were observed:
\begin{itemize}
    \item Generalization occurs long after training accuracy reached $100\%$. The jump in generalization is quite rapid and occurs after a large number of epochs (cf.~Fig.~\ref{fig:fig0}).
    \item There is a minimal amount of data (dependent on the task) that needs to be included into the training set in order for generalization to occur (cf.~Fig.~\ref{fig:fig3}b).
    \item Various forms of regularization improve how quickly grokking happens. Weight decay included in AdamW optimizer showed to be particularly effective (cf.~Fig.~\ref{fig:fig3}b).
\end{itemize}

In subsequent work \cite{liu2022towards},
the authors simplified the architecture to a single linear learnable encoder followed by a multilayer perceptron (MLP) decoder and showed that, even if the task is recast as a classification problem, grokking persists. They also interpret grokking as a competition between encoder and decoder, and developed a toy model of grokking as dynamics of the embeddings only. This model indeed leads to some quantitative predictions such as the critical amount of data needed for grokking to happen relatively fast.

\begin{figure}[!h]
    \centering
    \includegraphics[width=\textwidth]{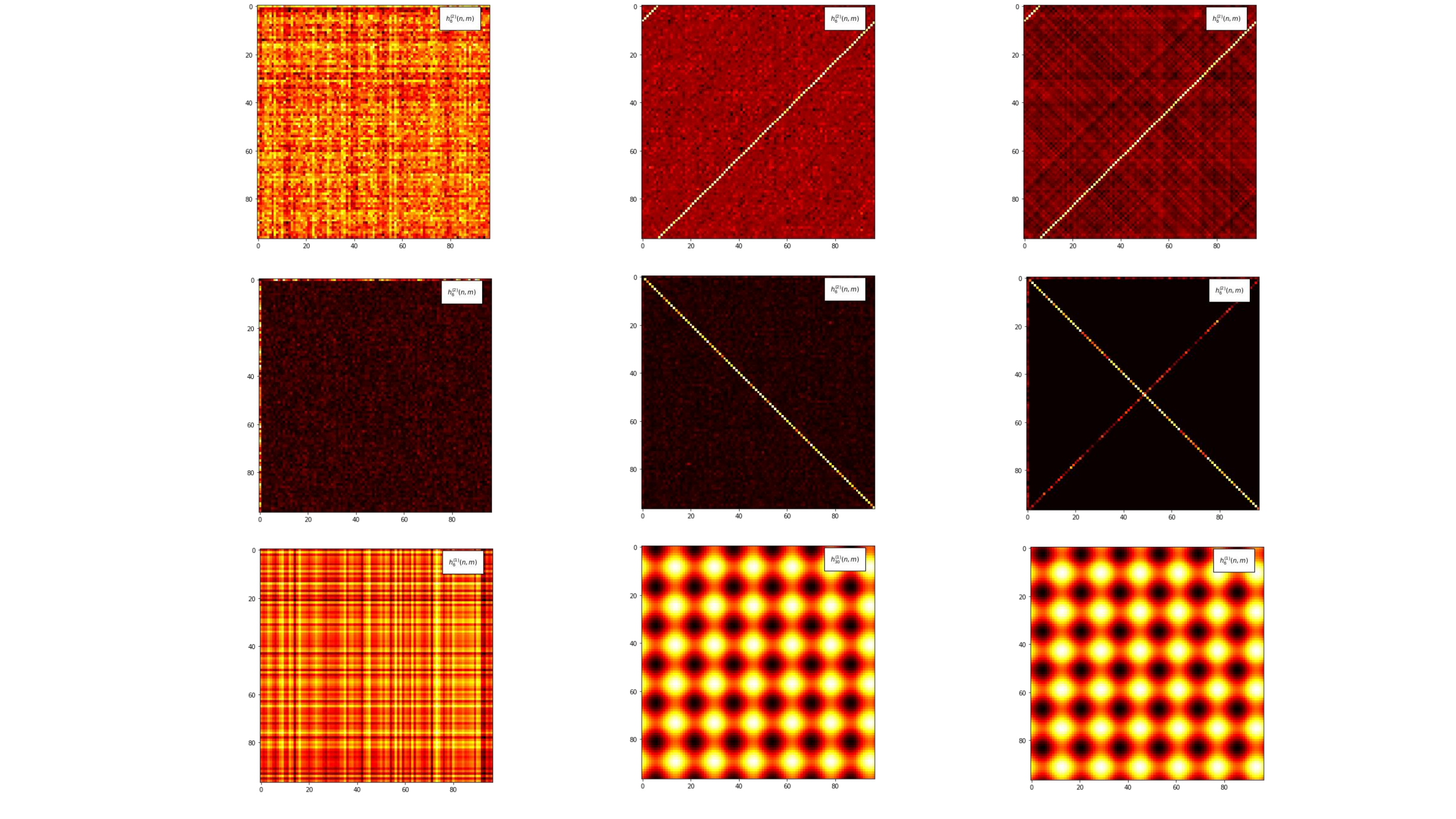}
    \caption{Preactivations. \textbf{First row}: Preactivation $ h^{(2)}_6(n,m)$. \textbf{Second row}: Fourier image of the Preactivation $h^{(2)}_6(n,m)$. \textbf{Third row}: Preactivation $h^{(1)}_6(n,m)$ or $h^{(1)}_{30}(n,m)$. \textbf{First column}: At initialization. \textbf{Second column}: Found by vanilla GD. The Fourier image shows a single series of peaks corresponding to $m+n = 6 \,\,\textrm{mod}\,\, 97$. \textbf{Third column}: Evaluated using the analytic solution \eqref{eq_sol1}-\eqref{eq_sol2}. The Fourier image shows the same peak as found by GD, but also weak peaks corresponding to $2m = 6 \,\,\textrm{mod}\,\, 97$, $2n = 6 \,\,\textrm{mod}\,\, 97$ and $m - n = 6 \,\,\textrm{mod}\,\, 97$ that were suppresed by the choice of phases via \eqref{eq_phases}.}
    \label{fig:fig1}
\end{figure}

In more recent work~\cite{thilak2022slingshot}, it was argued that if the Adam optimizer is used, then in order for grokking to happen without regularization, the training dynamics must undergo a slingshot -- a sudden explosion in the training loss -- which was followed by the rise of generalization. It was further shown that those slingshots and grokking can be turned on and off by tuning the $\epsilon$ parameter of the Adam optimizer.

In a blogpost~\cite{nanda2022mechanistic}, it was argued that the algorithm for modular addition learnt by a single-layer transformer can be reverse-engineered and is human-interpretable. It was further argued that (i) regularization is required for grokking and (ii) there should be no grokking in the infinite-data regime. Furthermore, many other algorithmic datasets were considered.

On a theoretical front, the authors of \cite{barak2022hidden} studied \emph{online} learning of the $(k,n)$ sparse parity problem where the network function is asked to compute  parity of $k$ bits in a length-$n$ string of random bits. In particular, they observed grokking both in under- and over-parametrized regimes. For large minibatch sizes, generalization was attributed to amplification of the information already present in the initial gradient (called Fourier gap) rather than to the diffusive search by 
stochastic gradient descent, and derived the scaling of grokking time with $n,k$ to be $n^{O(k)}$. 

Finally, \cite{liu2022omnigrok} studied grokking for non-algorithmic datasets and its dependence on the initialization, while \cite{vzunkovivc2022grokking} developed a solvable model of grokking in the teacher-student setup. 

To summarize, the available results, although undoubtedly inspiring, leave grokking on algorithmic datasets as a somewhat mysterious effect. Furthermore, the empirical results suggest that grokking provides a fascinating platform for quantitatively studying many fundamental questions of deep learning in a controlled setting. These include: (i) the precise role of regularization in deep nonlinear neural networks; (ii) feature learning; (iii) the role of training data distributions in optimization dynamics and generalization performance of the network; (iv) data-, parameter- and compute-efficiency of training; (v) interpretability of learnt features; and (vi) expressivity of architectures and complexity of tasks.

This motivates the present study, proposing and analyzing a minimal yet realistic
model and optimization process that lead to grokking on modular arithmetic tasks.

\section{Set up and overview of results}

In this Section we describe a very simple, solvable, setting where grokking takes place and learnt features can be understood analytically. We consider a two-layer MLP network without biases, given by
\begin{eqnarray} \label{eq_h1}
    && h^{(1)}_k(x) = \sqrt{\frac{1}{D}} \sum_{j=1}^{D} W^{(1)}_{kj} x_j\,, \qquad  z^{(1)}_i(x) = \phi(h^{(1)}_i(x))\,,
    \\\label{eq_h2}
    && h^{(2)}_q(x) = \frac{1}{N} \sum_{k=1}^N W^{(2)}_{qk} z^{(1)}_k(x)\,,
\end{eqnarray}
where $N$ is the width of the hidden layer, $D$ is the input dimension, and $\phi$ is an activation function.
At initialization the weights are sampled from the standard normal distribution $W^{(1)}, W^{(2)} \sim \mathcal N(0,1)$.
In Eqs.~\eqref{eq_h1}--\eqref{eq_h2}, we have chosen to follow the mean-field parametrization \cite{song2018mean}: this parametrization ensures that the analytic solution presented in the next Section remains finite in the large-$N$ limit\footnote{In the limit of infinite width, the meanfield parametrization allows for feature learning.}. 

Given this architecture, we then set up modular arithmetic tasks as classification problems. To this end, we fix an integer $p$ (that does not have to be prime) and consider functions over $\mathbb Z_p$. Each input integer is encoded as a one-hot vector. The output integer is also encoded as a one-hot vector. For the task of learning bivariate functions over  $\mathbb Z_p$ the input dimension is $2p$, the output dimension is $p$, the total number of points in the dataset is $p^2$, while the model~\eqref{eq_h1}--\eqref{eq_h2} has $3 N p$ parameters.
Finally, we split the dataset $\mathcal D$ into train $\mathcal D_{\rm train}$ and test $\mathcal D_{\rm test}$ subsets, and furnish this setup with the MSE loss function.\footnote{CSE loss can be used, if desired.}

Under this minimal setting grokking occurs consistently for many modular functions, provided enough epochs of training have taken place \textit{and} the fraction of data used for training, 
\begin{equation}
   \alpha \equiv \frac{|\mathcal D_{\rm train}|}{|\mathcal D|}\,, 
\end{equation}
is sufficiently large (if $\alpha$ is too small, generalization is not possible even after long training time). By adjusting width $N$, at fixed $\alpha$, we can tune between underparametrized and overparametrized regimes.   
The `simplest' optimizer that leads to grokking is the full-batch gradient descent. \emph{No explicit regularization is necessary for grokking to occur}. We have tried other optimizers and regularization methods such as AdamW, GD with weight decay and momentum, SGD with Batchnorm, and GD with Dropout. Generally, regularization and the use of adaptive optimizers produce two effects: (i) grokking happens after a smaller number of epochs and (ii) grokking happens at smaller $\alpha$. See Fig.~\ref{fig:fig3}.

In passing, we note that, in the case of quadratic activation the full network function takes an even simpler form
\begin{equation} \label{eq_f}
    f(x) = \frac{1}{D N}  W^{(2)} \left(W^{(1)} x\right)^2\,.
\end{equation}
This function is \emph{cubic} in parameters and \emph{quadratic} in its inputs. Eq.~\eqref{eq_f} is the simplest possible \emph{nonlinear} generalization of the `$u$-$v$' model studied in \cite{lewkowycz2020large}.
The exact results are derived for this particular choice (and can be generalized to other monomials if wished) while empirical results are only mildly sensitive to the choice of activation function.

Whether grokking happens or not depends on the modular function at hand assuming the architecture and optimizer are fixed. We show that for any function of the form $f(n,m) = f_1(n) + f_2(m)\,\, \textrm{mod}\,\,p$ as well as $\tilde{f}(n,m) = F(f_1(n) + f_2(m))\,\, \textrm{mod}\,\,p$ one can present an analytic solution for the weights that yield $100\%$ accuracy and these weights are approximately found by various optimizers with and without regularization. Functions of the form $g(n,m) = g_1(n) \cdot g_2(m) \,\, \textrm{mod}\,\,p$ can also be grokked, however we have failed to find the analytic expression for the weights. Functions of the form $f(n,m) + g(n,m) \,\, \textrm{mod} \,\, p$ are more difficult to grok: they require more epochs and larger $\alpha$.

In summary, our setup is simple enough to be analytically tractable but complex enough to exhibit representation learning and, consequently, grokking. 

\section{Interpretability: analytic expression for the weights}

%
\subsection{Modular addition}

In this Section we will exhibit the analytic expression for the weights that solve the modular addition task. Namely, the network supplied with these weights implements the following modular function
\begin{equation}
    f(n,m) = n + m \,\, \textrm{mod} \,\, p\,.
\end{equation}
This solution is approximate and can be made increasingly more accurate (meaning the test \emph{loss} can be made arbitrarily close to $0$) by increasing the width $N$. To simplify the presentation, we will discuss modular addition at length and then generalize the solution to a broad class of modular functions. In the next Section we will provide evidence that the GD and AdamW find the same solution. 

\textbf{Claim I.} If the network function has the form \eqref{eq_f} then the weights $W^{(1)}_{kn}$ and $W^{(2)}_{qk}$ solving the modular addition problem are given by
\begin{eqnarray} \label{eq_sol1}
    &&W^{(1)}_{kn} = 
    \begin{pmatrix}
    \cos \left( 2 \pi\frac{k}{p} n_1 +\varphi^{(1)}_k\right)
    \\
    \cos \left( 2 \pi\frac{k}{p} n_2 +\varphi^{(2)}_k\right)
    \end{pmatrix}^T\,, \qquad n=(n_1,n_2)
    \\ \label{eq_sol2}
    &&W^{(2)}_{qk} = \cos \left( -2 \pi\frac{k}{p} q  -\varphi^{(3)}_k\right)\,,
\end{eqnarray}
where we represent $W^{(1)}_{kn}$ as a row of two $N \times p$ matrices and $n_1, n_2 = 0, 1,\ldots,p-1$. The full size of $W^{(1)}_{kn}$ is $N \times 2p$. The phases $\varphi^{(1)}_k, \varphi^{(2)}_k$ and $\varphi^{(3)}_k$ are random, sampled from a uniform distribution and satisfy the constraint \eqref{eq_phases}.

\textbf{Reasoning. } Here we explain why and how the solution \eqref{eq_sol1}-\eqref{eq_sol2} works. There are two important ingredients in \eqref{eq_sol1}-\eqref{eq_sol2}. The first ingredient is the periodicity of weights with respect to the indices $n_1,n_2,q$. The set of frequencies is determined by the base of $\mathbb Z_p$. The full set of independent frequencies is obtained by varying $k$ from $0$ to $\frac{p-1}{2}$ if $p$ is odd and to $\frac{p}{2}$ if $p$ is even. The second ingredient is the set of phases $\varphi^{(1)}_k, \varphi^{(2)}_k, \varphi^{(3)}_k$. Indeed, Eqs. \eqref{eq_sol1}-\eqref{eq_sol2} solve modular addition \emph{only} after these phases are chosen appropriately. We will discuss the choice shortly.

To show that \eqref{eq_sol1}-\eqref{eq_sol2} solve modular addition we will perform the inference step analytically. Consider a general input $(n,m)$ represented as a pair of one-hot vectors stacked into a single vector of size $2p \times 1$.

The preactivations in the first layer are given by (we drop the normalization factors)
\begin{equation}
    h^{(1)}_k(n,m) = \cos \left( 2 \pi\frac{k}{p} n +\varphi^{(1)}_k\right) + \cos \left( 2 \pi\frac{k}{p} m +\varphi^{(2)}_k\right)\,.
\end{equation}
The activations in the first layer are given by 
\begin{equation}
    z^{(1)}_k(n,m) = \left(\cos \left( 2 \pi\frac{k}{p} n +\varphi^{(1)}_k\right) + \cos \left( 2 \pi\frac{k}{p} m +\varphi^{(2)}_k\right)\right)^2\,,
\end{equation}
which, after some trigonometry, becomes
\begin{eqnarray} \nonumber
    z^{(1)}_k(n,m) &=& 1 + \frac{1}{2}\left( \cos \left( 2 \pi\frac{k}{p} 2n +2\varphi^{(1)}_k\right) + \cos \left( 2 \pi\frac{k}{p} 2m +2\varphi^{(2)}_k\right)\right) 
    \\
    &+& \cos \left( 2 \pi\frac{k}{p} (n+m) +\varphi^{(1)}_k + \varphi^{(2)}_k\right) + \cos \left( 2 \pi\frac{k}{p} (n-m) +\varphi^{(1)}_k - \varphi^{(2)}_k\right)\,.
\end{eqnarray}
Finally, the preactivations in the second layer take form
\begin{eqnarray} \nonumber
h^{(2)}_q(n,m) &=& \frac{1}{4} \sum_{k=1}^N \cos \left( 2 \pi\frac{k}{p} (2n-q) +2\varphi^{(1)}_k - \varphi^{(3)}_k\right) + \cos \left( 2 \pi\frac{k}{p} (2n+q) +2\varphi^{(1)}_k + \varphi^{(3)}_k\right)
\\ \nonumber
&+& \frac{1}{4} \sum_{k=1}^N \cos \left( 2 \pi\frac{k}{p} (2m-q) +2\varphi^{(1)}_k - \varphi^{(3)}_k\right) + \cos \left( 2 \pi\frac{k}{p} (2m+q) +2\varphi^{(1)}_k + \varphi^{(3)}_k\right)
\\\nonumber
&+& \frac{1}{2} \sum_{k=1}^N \cos \left( 2 \pi\frac{k}{p} (n+m-q) +\varphi^{(1)}_k + \varphi^{(2)}_k - \varphi^{(3)}_k\right)
\\\nonumber
&+& \frac{1}{2} \sum_{k=1}^N \cos \left( 2 \pi\frac{k}{p} (n+m+q) +\varphi^{(1)}_k + \varphi^{(2)}_k + \varphi^{(3)}_k\right)
\\\nonumber
&+& \frac{1}{2} \sum_{k=1}^N \cos \left( 2 \pi\frac{k}{p} (n-m-q) +\varphi^{(1)}_k - \varphi^{(2)}_k - \varphi^{(3)}_k\right)
\\\nonumber
&+& \frac{1}{2} \sum_{k=1}^N \cos \left( 2 \pi\frac{k}{p} (n-m+q) +\varphi^{(1)}_k - \varphi^{(2)}_k + \varphi^{(3)}_k\right)
\\ \label{eq_omg}
&+& \sum_{k=1}^N \cos \left( 2 \pi\frac{k}{p} q + \varphi^{(3)}_k\right)\,.
\end{eqnarray}

Expression \eqref{eq_omg} does not yet perform modular addition. Observe that each term in \eqref{eq_omg} is a sum of waves with different phases, but systematically ordered frequencies. We are going to choose the phases $\varphi^{(1)}_k, \varphi^{(2)}_k, \varphi^{(3)}_k$ to ensure constructive interference in the third line of \eqref{eq_omg}. The simplest choice is to take
\begin{equation} \label{eq_phases}
    \varphi^{(1)}_k + \varphi^{(2)}_k = \varphi^{(3)}_k\,.
\end{equation}
Then the term in the third line of \eqref{eq_omg} takes form 
\begin{equation}
    \frac{1}{2} \sum_{k=1}^N \cos \left( 2 \pi\frac{k}{p} (n+m-q)\right) = \frac{N}{2}\delta(n+m-q)\,,
\end{equation}
where $\delta(n+m-q)$ is the modular version of the $\delta$-function. It is equal to $1$ when $n+m-q = 0 \,\, \textrm{mod}\,\, p$ and is equal to $0$ otherwise. This concludes the constructive part of the interference. 

Next, we need to ensure that all other waves (\emph{i.e.} all terms, but the third term in \eqref{eq_omg}) interfere destructively. Fortunately, this can be accomplished by observing that the constraint \eqref{eq_phases} leaves some phases in every single term in \eqref{eq_omg} apart from the third one. We will spare the reader the explicit expression. Every remaining term takes form
\begin{equation}
    \frac{1}{2} \sum_{k=1}^N \cos \left( 2 \pi\frac{k}{p} s +\varphi_k\right)\,,
\end{equation}
where $s$ is an integer and $\varphi_k$ is a linear combination of $\varphi^{(1)}_k$ and $\varphi^{(2)}_k$. We now assume that $\varphi^{(1)}_k$ and $\varphi^{(2)}_k$ are uniformly distributed random numbers. Then so are $\varphi_k$. For any appreciable $N$ (see Fig.~\ref{fig:fig3}b) we have 
\begin{equation} \label{eq_interference}
    \sum_{k=1}^N \cos \left( 2 \pi\frac{k}{p} s +\varphi_k\right) \ll N\,,
\end{equation}
which implies that every term in \eqref{eq_omg} can be neglected compared to the third term.
Thus, for reasonable values of $N$ (and restoring normalisation) the network function $h^{(2)}_q(n,m)$ takes form  
\begin{equation}
    h^{(2)}_q(n,m) \approx \frac{1}{2} \sum_{k=1}^N \cos \left( 2 \pi\frac{k}{p} (n+m-q)\right) = \frac{1}{2D}\delta(n+m-q)\,.
\end{equation}
In the limit of large $N$ the approximation becomes increasingly more accurate. Note that $h^{(2)}_q(n,m)$ is finite in the infinite width limit.

The test accuracy of the solution \eqref{eq_sol1}-\eqref{eq_sol2} \emph{increases with width}. For larger $N$ the interference is stronger leading to the better approximation of the $\delta$-function and, ultimately, to better accuracy. In this example we clearly see that larger width does \emph{not} imply a larger number of relevant features. Instead, it introduces redundancy: each frequency appears several times with different random phases ultimately leading to a better wave interference.

We emphasize that, the weights \eqref{eq_sol1}-\eqref{eq_sol2} are not iid. At fixed $k$ the weights $W^{(1)}_{kn}, W^{(2)}_{qk}$ are strongly correlated with each other. This provides a non-trivial yet analytically tractable example of a correlated, non-linear, network far away from the Gaussian limit.

\begin{figure}[!h]
    \centering
    \includegraphics[width=\textwidth]{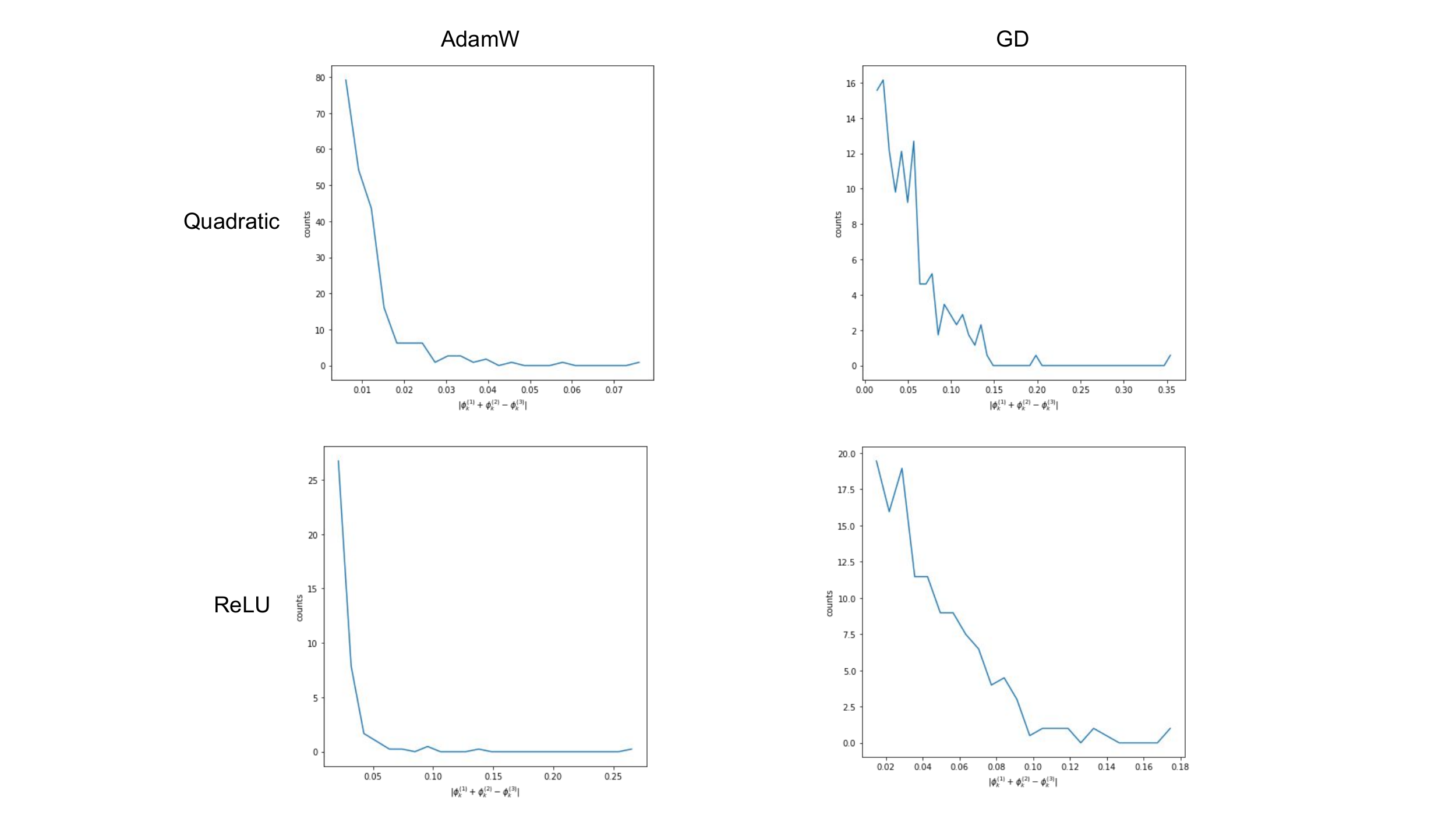}
    \caption{Solutions found by the optimizer. In all cases distribution of $\varphi^{(1)}_k + \varphi^{(2)}_k - \varphi^{(3)}_k$ is strongly peaked around $0$. The solutions found by AdamW are closer to the analytic ones because the phases are peaked stronger around $0$. Note that for solutions found by the optimizer the phases are not iid which leads to the better accuracy.}
    \label{fig:fig2}
\end{figure}

The weights \eqref{eq_sol1}-\eqref{eq_sol2} also work for other activation functions, including ReLU, however the $100\%$ accuracy is achieved at higher width compared to quadratic activation function (more details in Appendix B).

\subsection{General modular functions and complexity}

The solution \eqref{eq_sol1}-\eqref{eq_sol2} can be easily generalized to represent a general modular function of the form
\begin{equation}
    f(n,m) = f_1(n) + f_2(m) \,\, \textrm{mod}\,\,p \,,
\end{equation}
where $f_1, f_2$ are arbitrary modular functions of a single variable. The generalization becomes obvious once we observe that the proof presented in Section~$3$ holds verbatim upon replacing $n\rightarrow f_1(n)$ and $m\rightarrow f_2(m)$ leading to a $\delta$-function supported on $f_1(n) + f_2(m) - q = 0 \,\, \textrm{mod}\,\, p$.
These solutions are also found by the optimizer just like in the case of modular addition. More precisely, we claim

\textbf{Claim II. }If the network function has the form \eqref{eq_f} then the weights $W^{(1)}_{kn}$ and $W^{(2)}_{qk}$ solving the modular task $f(n,m) = f_1(n) + f_2(m) \,\, \textrm{mod}\,\,p$ are given by
\begin{equation} \label{eq_solf1}
    W^{(1)}_{kn} = 
    \begin{pmatrix}
    \cos \left( 2 \pi\frac{k}{p} f_1(n_1) +\varphi^{(1)}_k\right)
    \\
    \cos \left( 2 \pi\frac{k}{p} f_2(n_2) +\varphi^{(2)}_k\right)
    \end{pmatrix}\,, \qquad n=(n_1,n_2)
\end{equation}
and Eq. \eqref{eq_sol2}. The weights depend on the modular arithmetic task at hand.  Furthermore, for this class of tasks the weights in the readout layer are unchanged. A simple example is $f(n,m) = n^2 + m^2$. The activations for this task are presented in the Appendix C. 

\textbf{Corollary. }Given the Claim II, a more general modular task $\tilde f(n,m) = F(f_1(n) + f_2(m))\,\, \text{mod}\,\,p$, can be solved, assuming that $F$ is invertible. This is accomplished by modifying the readout layer weights as follows
\begin{equation}\label{eq_solfF2}
    W^{(2)}_{qk} = \cos \left( -2 \pi\frac{k}{p} F^{-1}(q)  -\varphi^{(3)}_k\right)\,.
\end{equation}
This solution approximates $\delta( f_1(n) + f_2(m) - F^{-1}(q))$, which is equivalent to the $\delta$-function supported on the claimed modular task $\delta( F(f_1(n) + f_2(m)) - q)$ assuming $F^{-1}$ is single-valued. Note that application of $F^{-1}$ must follow modular arithmetic rules. If $F^{-1}$ is not single-value then the accuracy will be approximately $100\% / b$, where $b$ is the number of branches. A simple example is $f(n,m) = (n + m)^2$. The activations for this task are presented in the Appendix. Analytic solution has accuracy $\approx 50\%$ since $F^{-1}(x) = x^{\frac{1}{2}} \,\, \text{mod}\,\,p$, which has two branches.

The architecture \eqref{eq_f} can also learn modular multiplication, however we do not posses an analytic solution for that case.

Broadly speaking, a bivariate modular function is a $p \times p$ table where each entry can take values between $0$ and $p-1$. There are $p^{p^2}$ such tables. Clearly, grokking is not possible on the overwhelming majority of such functions, because this set includes placing random integers in each entry of the table. Some modular functions, namely the ones that involve addition \emph{and} multiplication, \emph{and} are not of the form $\tilde f$ are substantially harder to learn. They require more data, more time and do not always yield $100\%$ test accuracy after grokking. One particularly interesting example was found by \cite{power2022grokking}, $f(n,m) = n^3 + nm^2 + m$, which does not generalize even for $\alpha>0.9$, both for transformer and MLP architectures. Some examples are discussed in Appendix. It is not clear how to predict which functions will generalize and which will not given an architecture.

\section{Properties of solutions found by gradient descent}

%
\subsection{General properties}
In this Section we show that optimization of the network \eqref{eq_h1}-\eqref{eq_h2} yields a solution that is very close to the one we proposed in the previous Section.

\begin{figure}[!h]
    \centering
    \includegraphics[width=\textwidth]{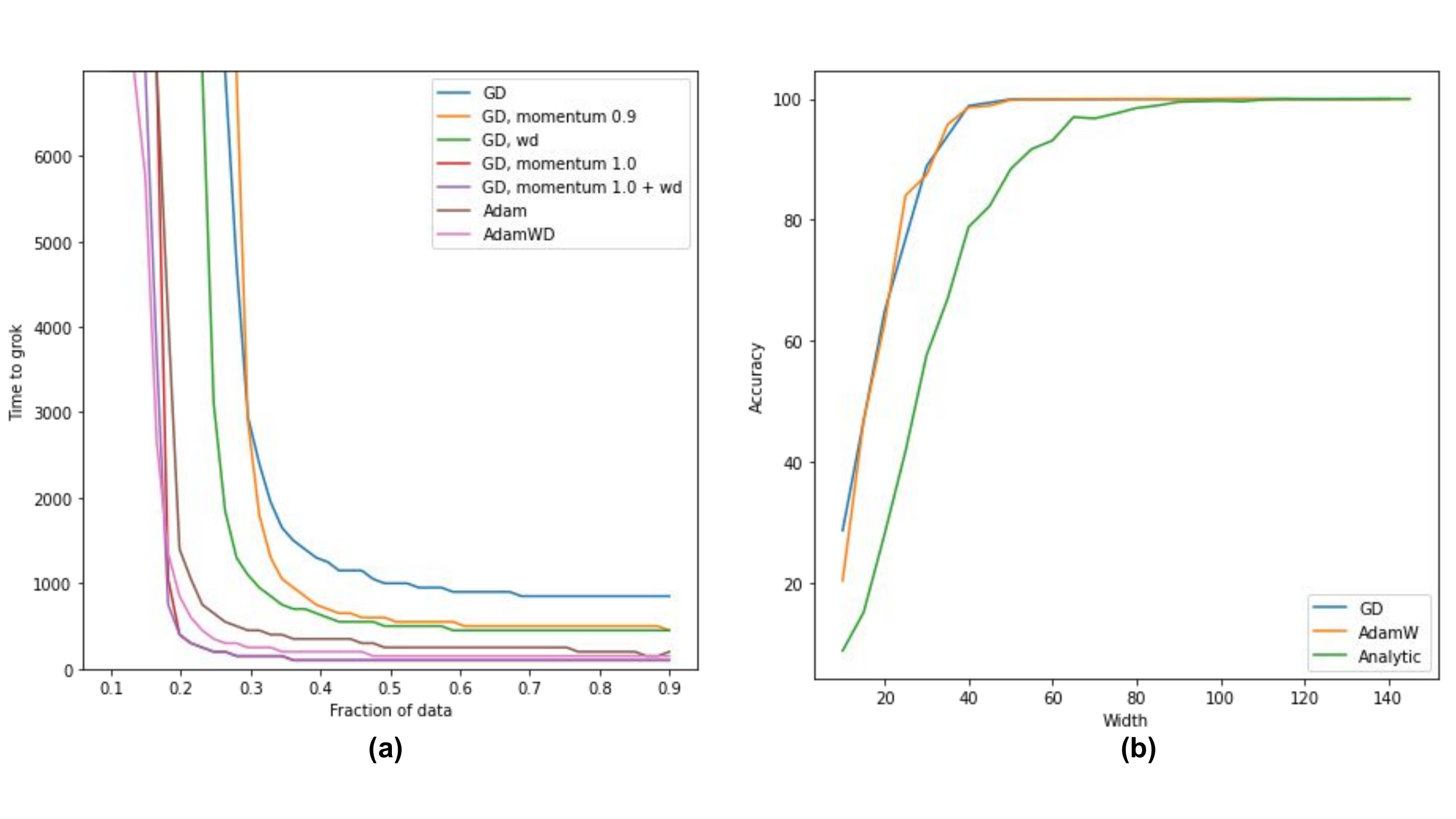}
    \caption{Scaling with width and data. \textbf{(a)} Grokking time vs. the amount of training data for various optimizers. The abrupt change in grokking time is observed at different $\alpha$. Momentum appears to play a major role both in reducing grokking time and $\alpha$. \textbf{(b)}: Test accuracy as a function of width for the solution found by GD, AdamW and for the analytic solution \eqref{eq_sol1}--\eqref{eq_sol2}. The optimizer can tune phases better than random uniform distribution in order to ensure better cancellations. The shape of the curves also depends on the amount of data used for training and number of epochs. Here we took $\alpha = 0.5$ and trained longer for GD.}
    \label{fig:fig3}
\end{figure}

As can be seen in Fig.~\ref{fig:fig0} during the optimization the network first overfits the train data. The periodic structure in weights and activations does not form at that point. Train loss slowly gets smaller until it either (i) saturates leading to a memorizing solution without grokking the problem, or (ii) after a period of slow decrease, it slightly accelerates. It is during that time grokking and feature formation take place.
The test loss is \emph{non-monotonic} and reaches a local maximum right before grokking happens. In the memorizing phase test loss never leaves this local maximum. This general behaviour appears to be insensitive to either optimizer used, loss function or modular function (\emph{i.e.} dataset) in question.

We then show empirically that independently of the optimizer and the loss function the features found by optimization in the grokking phase are indeed periodic functions with frequencies $\frac{2\pi k}{p}$ where $k=0,\ldots,p-1$. If the width is larger than $\frac{p-1}{2}$ then multiple copies of these functions are found with different phases. The phases are approximately random and satisfy the constraint \eqref{eq_phases} approximately as we show in Fig.\ref{fig:fig2}. Given the simplicity of the setup, the basic explanation for grokking must be quite banal. At some point in training, the only way to decrease training loss is to start learning the ``right'' features.

\subsection{Scaling}
Scaling with width and dataset size are presented on Fig.~\ref{fig:fig3}.
The accuracy of solution \eqref{eq_sol1}-\eqref{eq_sol2} favorably scales with width. This stems from the simple fact that destructive interference condition \eqref{eq_interference} becomes increasingly more accurate with larger $N$. The test accuracy of trained network also increases with the width, reaching perfect accuracy before the analytic solution does, which is not surprising because optimizer can tune the individual phases to ensure better performance.

The grokking time scales with the amount of data. Both for GD and AdamW there is a critical amount of data $\alpha_c$ such that grokking is possible. The precise value of $\alpha_c$ is hard to determine because of the long time scales needed for grokking close to $\alpha_c$. This is clearly seen on Fig.\ref{fig:fig3}. AdamW appears to be more data-efficient than GD, however it is difficult to rule out the possibility that for $\alpha\approx 0.2$ GD requires extremely long time scales to show grokking. The value of $\alpha_c$ also depends on how the training set is sampled. One can imagine a random sampling or a guided algorithmic choice of training examples. The latter will lead to smaller $\alpha_c$.

\subsection{Dynamics}
In this Section we introduce an empirical measure that quantifies the feature learning for the modular addition task. To define such measure we turn to the exact solution \eqref{eq_sol1}- \eqref{eq_sol2}. We will utilize the fact that periodic weights are peaked in Fourier space, while random weights are not.

To define the measure of feature learning, we first transform the weights $W^{(1)}_{nk}$ to a Fourier space with respect to index $n$. Denote the transformed weights $\tilde{W}^{(1)}_{\nu k}$. If the weights are periodic, then Fourier-transformed weights are \emph{localized} in $\nu$, \emph{i.e.} for most values of $\nu$ we have $\tilde{W}^{(1)}_{\nu k}\approx 0$ except for a few values determined by the frequency $\frac{2\pi}{p}k$. At initialization, when the weights are random the Fourier-transformed weights are \emph{delocalized}, \emph{i.e.} will take roughly equal values for any $\nu$.

We introduce a measure of localization known as the inverse participation ratio (IPR). It is routinely used in localization physics \cite{girvin2019modern} as well as network theory \cite{pastor2016distinct}. We define IPR in terms of the normalized Fourier-transformed weights
\begin{equation}
    \textrm{IPR}_r(k) = \sum_{\nu=1}^D |\tilde{w}^{(1)}_{\nu k}|^{2r}\,,\qquad \text{where}\qquad \tilde{w}^{(1)}_{\nu k} = \frac{\tilde{W}^{(1)}_{\nu k}}{\sqrt{\sum_{\nu=1}^D (\tilde{W}^{(1)}_{\nu k})^2}}\,,
\end{equation}
and $r$ is a parameter traditionally taken to be $2$. It follows from the definition that $\textrm{IPR}_1(k) = 1$ for any $k$.
Unfortunately, $\textrm{IPR}_r(k)$ is defined per neuron. We would like a single measure for all of the weights in a given layer. Thus, we introduce the average IPR
\begin{equation}
    \overline{\textrm{IPR}}_r = \frac{1}{N} \sum_{k=1}^N \textrm{IPR}_r(k)\,.
\end{equation}
Larger values of $\overline{\textrm{IPR}}_r$ indicate that the weights are more periodic, while the smaller values indicate that the weights are more random.

\begin{figure}[!h]
    \centering
    \includegraphics[width=\textwidth]{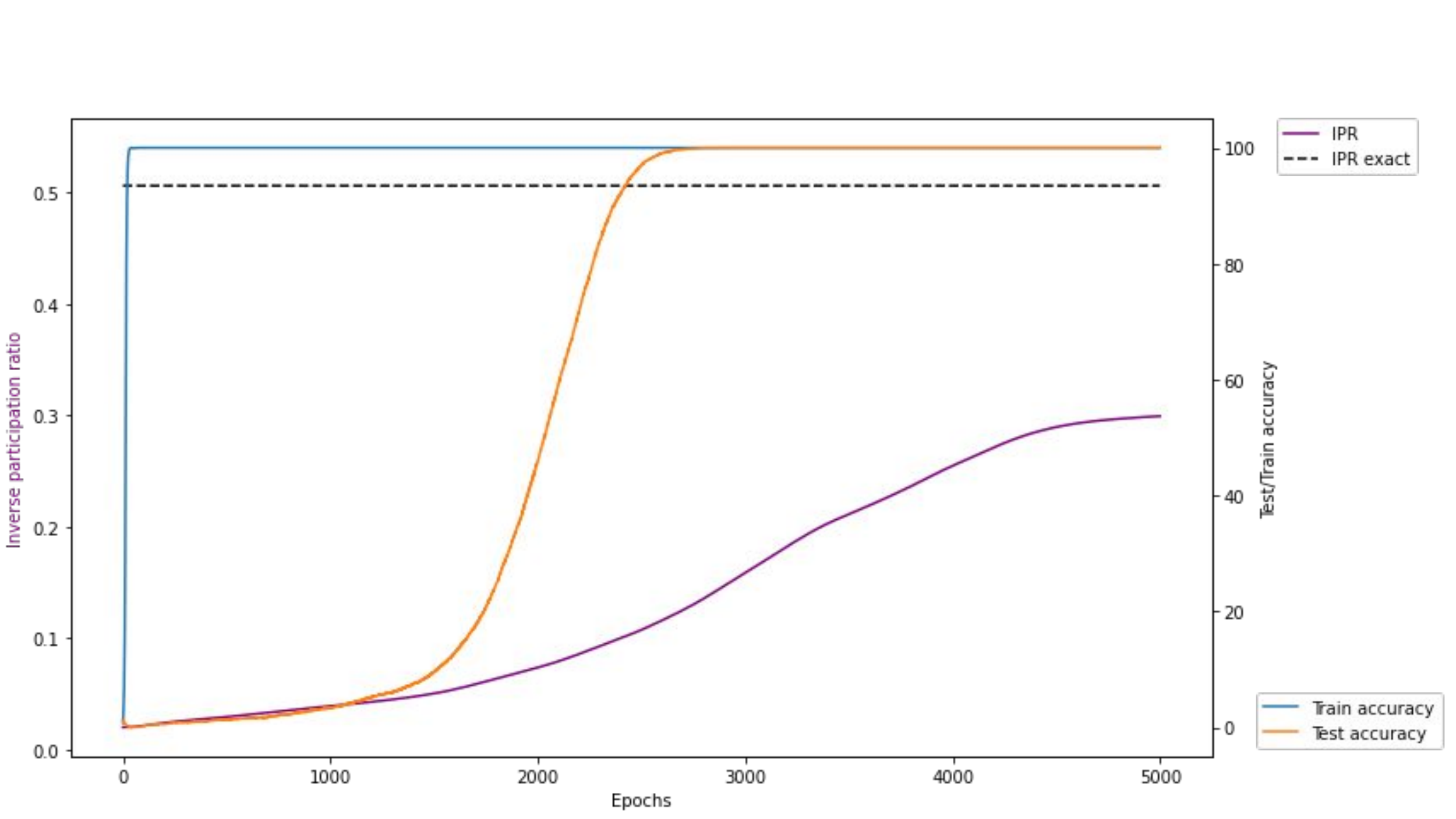}
    \caption{Inverse participation ratio. IPR plotted against the dynamics (under AdamW) of train and test accuracy. Empirically, we see $4$ regimes: (i) early training when IPR grows linearly and slowly; (ii) transition from slow liner growth to fast linear growth. This transition period coincides with grokking; (iii) fast linear growth, that starts after $100\%$ test accuracy was reached; (iv) saturation, once weights became periodic. The dashed line indicates $\overline{\textrm{IPR}}_2$ for the exact solution \eqref{eq_sol1}-\eqref{eq_sol2}. The gap between the two indicates that even in the final solution there is quite a bit of noise leading do some mild delocalization in Fourier space. More training and more data helps to reduce the gap.}
    \label{fig:fig4}
\end{figure}

We plot $\overline{\textrm{IPR}}_2$ as a function of time in Fig.~\ref{fig:fig4}. It is clear that there is an upward trend from the very beginning of training. Onset of grokking is correlated with the sharp increase of rate of IPR growth.

%
%


\section{Conclusions and discussions}

%
\subsection{Conclusions}
We have presented a simple architecture that exhibits grokking on a variety of modular arithmetic problems. The architecture \eqref{eq_f} is simple enough to determine the weights and features that solve modular addition problems analytically, leading to complete interpretability of what was learnt by the model: the network is learning a $\delta$-function represented by a complete set of trigonometric functions with frequencies determined by the base of modular addition; the phases are chosen to ensure that waves concentrated on $m+n = q \,\, \text{mod}\,\, p$ interfere constructively. 

As suggested in some literature before, we reiterate that grokking is likely to be intimately connected to feature learning.
In particular, random feature models such as infinitely-wide neural networks (in the NTK regime) do not exhibit grokking, at least on the tasks that involve modular functions.
In addition, Ref.~\cite{liu2022towards} argued that grokking is due to the competition between encoder and decoder. While it is certainly true in their model, in the present case there is no learnable encoder but grokking is still present. In our minimal setup, the simplest explanation for grokking is that once training loss reached a certain value, the only way to further minimize it is to start learning the right features.

\subsection{Discussions}
We close with a discussion of open problems and directions.

Different modular functions clearly fit into different complexity classes: (i) functions that can be learnt easily; (ii) functions that can be learnt with a lot of data and training time; and (iii) functions that cannot be learnt at all (at least within the class of architectures we and \cite{power2022grokking} have considered). It would be interesting to (1) define the notion of complexity rigorously as a computabe quantity and (2) construct architectures/optimizers that can learn more complex modular functions (or argue that it cannot be done).

A neural network can learn a smooth approximation to complicated modular operations, such as modular square root and modular logarithm. It would be interesting to determine if these approximations provide any practical gain over known algorithms that perform these operations as well as to enable the networks to operate over large numbers.

The critical amount of data needed for generalization, $\alpha_c$, is likely to be computable as well, and is a measure of complexity of a modular function. We would like to have an expression for the absolute minimal value of $\alpha_c$ (\emph{i.e.} minimized over all possible ML methods). This value is also an implicit function of modulus $p$, and the modular functions with larger modulus are likely simpler since we find empirically that $\alpha_c$ is a decreasing function of $p$. The value of $\alpha_c$ further depends on how training set is sampled from the entire dataset; the appropriate choice of the sampling method may thus improve the data efficiency.

While grokking happens in a very simple setting described here, adaptive methods and regularization improve both speed and data efficiency. It might be possible to characterize these improvements quantitatively.

Modular functions of many variables can be grokked as well and, in some cases, the corresponding analytic solution can be constructed. It is possible that the analytic solution can inform a type of architecture one should be using, e.g., in applications of deep learning to cryptography.

Presented solutions only work for a single-hidden-layer neural network. To quantify the role of depth, we would like to have examples of algorithmic tasks that require a deeper architecture. For instance, it is possible that deep convolutional architectures, given an appropriate algorithmic dataset with hierarchical structure, would admit a solution in terms of wavelets rather than Fourier components~\cite{cheng2021quantify}.

In real-world datasets and tasks that require feature learning, it is possible that grokking is happening but the jumps in generalization after learning a new feature may be so small that we perceive a continuous learning curve. To elucidate this point further, it is important to construct a realistic model of datasets and tasks with controllable amount of hierarchical structure.  More broadly, it would be very interesting to characterize grokking in terms that are not specific to a particular problem or a particular model and to establish whether it occurs in more traditional ML settings.

Given the simplicity of our model \eqref{eq_f}, loss function (MSE) and optimization algorithm (vanilla GD), it is plausible that some aspects of the training dynamics -- not just the solution at the end of training -- can be treated analytically.
As the training and test losses show peculiar dynamics, it would be interesting to understand the structure of the loss landscape to explain the dynamics, in particular what happens at the onset of generalization and why it is so abrupt. Perhaps methods described in \cite{roberts2021principles} -- where the feature kernel and the neural tangent kernel can be computed analytically throughout the training --  will take a particularly simple form in this setting. 

There are certainly many other directions that the reader may be interested in exploring.

\begin{ack}
 Discussions with N.~Ardalani, Y.~Bahri, M.~Barkeshli, L.~Bottou, T.~Can, F.~Charton,  D.~Doshi, S.~Ganguli, P.~Glorioso, B.~Hanin, T.~He, I.~Molybog, M.~Paul, D.~Roberts,  A.~Saxe, D.~Schwab and  S.~Yaida are acknowledged. I am particularly grateful to S.~Yaida, D.~Roberts, and B.~Hanin for encouraging, detailed and insightful feedback on the manuscript. A.G.’s work at the University of Maryland was supported in part by NSF CAREER Award DMR-2045181, Sloan Foundation and the Laboratory for Physical Sciences through the Condensed Matter Theory Center.
\end{ack}

{
\small
\nocite{*}
\bibliography{Bibliography}

\begin{thebibliography}{16}
\providecommand{\natexlab}[1]{#1}
\providecommand{\url}[1]{\texttt{#1}}
\expandafter\ifx\csname urlstyle\endcsname\relax
  \providecommand{\doi}[1]{doi: #1}\else
  \providecommand{\doi}{doi: \begingroup \urlstyle{rm}\Url}\fi

\bibitem[Barak et~al.(2022)Barak, Edelman, Goel, Kakade, Malach, and
  Zhang]{barak2022hidden}
Boaz Barak, Benjamin~L Edelman, Surbhi Goel, Sham Kakade, Eran Malach, and
  Cyril Zhang.
\newblock Hidden progress in deep learning: Sgd learns parities near the
  computational limit.
\newblock \emph{arXiv preprint arXiv:2207.08799}, 2022.

\bibitem[Cheng and M{\'e}nard(2021)]{cheng2021quantify}
Sihao Cheng and Brice M{\'e}nard.
\newblock How to quantify fields or textures? a guide to the scattering
  transform.
\newblock \emph{arXiv preprint arXiv:2112.01288}, 2021.

\bibitem[Girvin and Yang(2019)]{girvin2019modern}
Steven~M Girvin and Kun Yang.
\newblock \emph{Modern condensed matter physics}.
\newblock Cambridge University Press, 2019.

\bibitem[Jacot et~al.(2018)Jacot, Gabriel, and Hongler]{jacot2018neural}
Arthur Jacot, Franck Gabriel, and Cl{\'e}ment Hongler.
\newblock Neural tangent kernel: Convergence and generalization in neural
  networks.
\newblock \emph{Advances in neural information processing systems}, 31, 2018.

\bibitem[Lee et~al.(2019)Lee, Xiao, Schoenholz, Bahri, Novak, Sohl-Dickstein,
  and Pennington]{lee2019wide}
Jaehoon Lee, Lechao Xiao, Samuel Schoenholz, Yasaman Bahri, Roman Novak, Jascha
  Sohl-Dickstein, and Jeffrey Pennington.
\newblock Wide neural networks of any depth evolve as linear models under
  gradient descent.
\newblock \emph{Advances in neural information processing systems}, 32, 2019.

\bibitem[Lewkowycz et~al.(2020)Lewkowycz, Bahri, Dyer, Sohl-Dickstein, and
  Gur-Ari]{lewkowycz2020large}
Aitor Lewkowycz, Yasaman Bahri, Ethan Dyer, Jascha Sohl-Dickstein, and Guy
  Gur-Ari.
\newblock The large learning rate phase of deep learning: the catapult
  mechanism.
\newblock \emph{arXiv preprint arXiv:2003.02218}, 2020.

\bibitem[Liu et~al.(2022{\natexlab{a}})Liu, Kitouni, Nolte, Michaud, Tegmark,
  and Williams]{liu2022towards}
Ziming Liu, Ouail Kitouni, Niklas Nolte, Eric~J Michaud, Max Tegmark, and Mike
  Williams.
\newblock Towards understanding grokking: An effective theory of representation
  learning.
\newblock \emph{arXiv preprint arXiv:2205.10343}, 2022{\natexlab{a}}.

\bibitem[Liu et~al.(2022{\natexlab{b}})Liu, Michaud, and
  Tegmark]{liu2022omnigrok}
Ziming Liu, Eric~J Michaud, and Max Tegmark.
\newblock Omnigrok: Grokking beyond algorithmic data.
\newblock \emph{arXiv preprint arXiv:2210.01117}, 2022{\natexlab{b}}.

\bibitem[Nanda and Lieberum(2022)]{nanda2022mechanistic}
Neel Nanda and Tom Lieberum.
\newblock A mechanistic interpretability analysis of grokking.
\newblock \emph{Alignment Forum}, Aug 2022.
\newblock URL
  \url{https://www.alignmentforum.org/posts/N6WM6hs7RQMKDhYjB/a-mechanistic-interpretability-analysis-of-grokking}.

\bibitem[Pastor-Satorras and Castellano(2016)]{pastor2016distinct}
Romualdo Pastor-Satorras and Claudio Castellano.
\newblock Distinct types of eigenvector localization in networks.
\newblock \emph{Scientific reports}, 6\penalty0 (1):\penalty0 1--9, 2016.

\bibitem[Power et~al.(2022)Power, Burda, Edwards, Babuschkin, and
  Misra]{power2022grokking}
Alethea Power, Yuri Burda, Harri Edwards, Igor Babuschkin, and Vedant Misra.
\newblock Grokking: Generalization beyond overfitting on small algorithmic
  datasets.
\newblock \emph{arXiv preprint arXiv:2201.02177}, 2022.

\bibitem[Roberts et~al.(2021)Roberts, Yaida, and Hanin]{roberts2021principles}
Daniel~A Roberts, Sho Yaida, and Boris Hanin.
\newblock The principles of deep learning theory.
\newblock \emph{arXiv preprint arXiv:2106.10165}, 2021.

\bibitem[Song et~al.(2018)Song, Montanari, and Nguyen]{song2018mean}
Mei Song, Andrea Montanari, and P~Nguyen.
\newblock A mean field view of the landscape of two-layers neural networks.
\newblock \emph{Proceedings of the National Academy of Sciences}, 115\penalty0
  (33):\penalty0 E7665--E7671, 2018.

\bibitem[Thilak et~al.(2022)Thilak, Littwin, Zhai, Saremi, Paiss, and
  Susskind]{thilak2022slingshot}
Vimal Thilak, Etai Littwin, Shuangfei Zhai, Omid Saremi, Roni Paiss, and Joshua
  Susskind.
\newblock The slingshot mechanism: An empirical study of adaptive optimizers
  and the grokking phenomenon.
\newblock \emph{arXiv preprint arXiv:2206.04817}, 2022.

\bibitem[Yang and Hu(2020)]{yang2020feature}
Greg Yang and Edward~J Hu.
\newblock Feature learning in infinite-width neural networks.
\newblock \emph{arXiv preprint arXiv:2011.14522}, 2020.

\bibitem[{\v{Z}}unkovi{\v{c}} and Ilievski(2022)]{vzunkovivc2022grokking}
Bojan {\v{Z}}unkovi{\v{c}} and Enej Ilievski.
\newblock Grokking phase transitions in learning local rules with gradient
  descent.
\newblock \emph{arXiv preprint arXiv:2210.15435}, 2022.

\end{thebibliography}
\bibliographystyle{plainnat}
}


\newpage
\appendix

\section{Complex network}

A simpler network that solves modular addition problem can be phrased using complex weights. This structure would also be more friendly to physicists. The complex solution takes form
\begin{eqnarray} \label{eq_sol1a}
    &&W^{(1)}_{kn} = 
    \begin{pmatrix}
    e^{ 2 \pi i\frac{k}{p} n_1 +i\varphi^{(1)}_k}
    \\
    e^{2 \pi i\frac{k}{p} n_2 +i\varphi^{(2)}_k}
    \end{pmatrix}\,, \qquad n=(n_1,n_2)
    \\ \label{eq_sol2a}
    &&W^{(2)}_{qk} = e^{ -2 \pi i\frac{k}{p} q  -i\varphi^{(3)}_k}\,,
\end{eqnarray}

We can take quadratic activation function that simply squares the preactivations. The first preactivation and activation are given by
\begin{eqnarray}
    && h^{(1)}(n,m) = e^{ 2 \pi i\frac{k}{p} n +i\varphi^{(1)}_k} + e^{ 2 \pi i\frac{k}{p} m +i\varphi^{(2)}_k}\,,
    \\
    && z^{(1)}(n,m) = e^{ 2 \pi i\frac{k}{p} 2n +i\varphi^{(1)}_k} + e^{ 2 \pi i\frac{k}{p} 2m +i\varphi^{(2)}_k} + 2 e^{ 2 \pi i\frac{k}{p} (n+m) +i(\varphi^{(1)}_k+\varphi^{(2)}_k)}\,.
\end{eqnarray}
The final activation is given by
\begin{eqnarray}
    h^{(2)}(n,m) &=& \sum_{k=1}^N \left(e^{ 2 \pi i\frac{k}{p} (2n-q) +i(\varphi^{(1)}_k-\varphi^{(3)}_k)} + e^{ 2 \pi i\frac{k}{p} (2m-q) +i(\varphi^{(2)}_k-\varphi^{(3)}_k)}\right.
    \\
    &+& \left.2 e^{ 2 \pi i\frac{k}{p} (n+m-q) +i(\varphi^{(1)}_k+\varphi^{(2)}_k-\varphi^{(3)}_k)}\right)\,.
\end{eqnarray}
Similarly setting 
\begin{equation}
    \varphi^{(1)}_k+\varphi^{(2)}_k-\varphi^{(3)}_k = 0
\end{equation}
yields the constructive interference for the output supported on $(n+m-q) = 0 \,\, \text{mod}\,\,p$.

\section{Other activations}

\begin{figure}[!h]
    \centering
    \includegraphics[width=\textwidth]{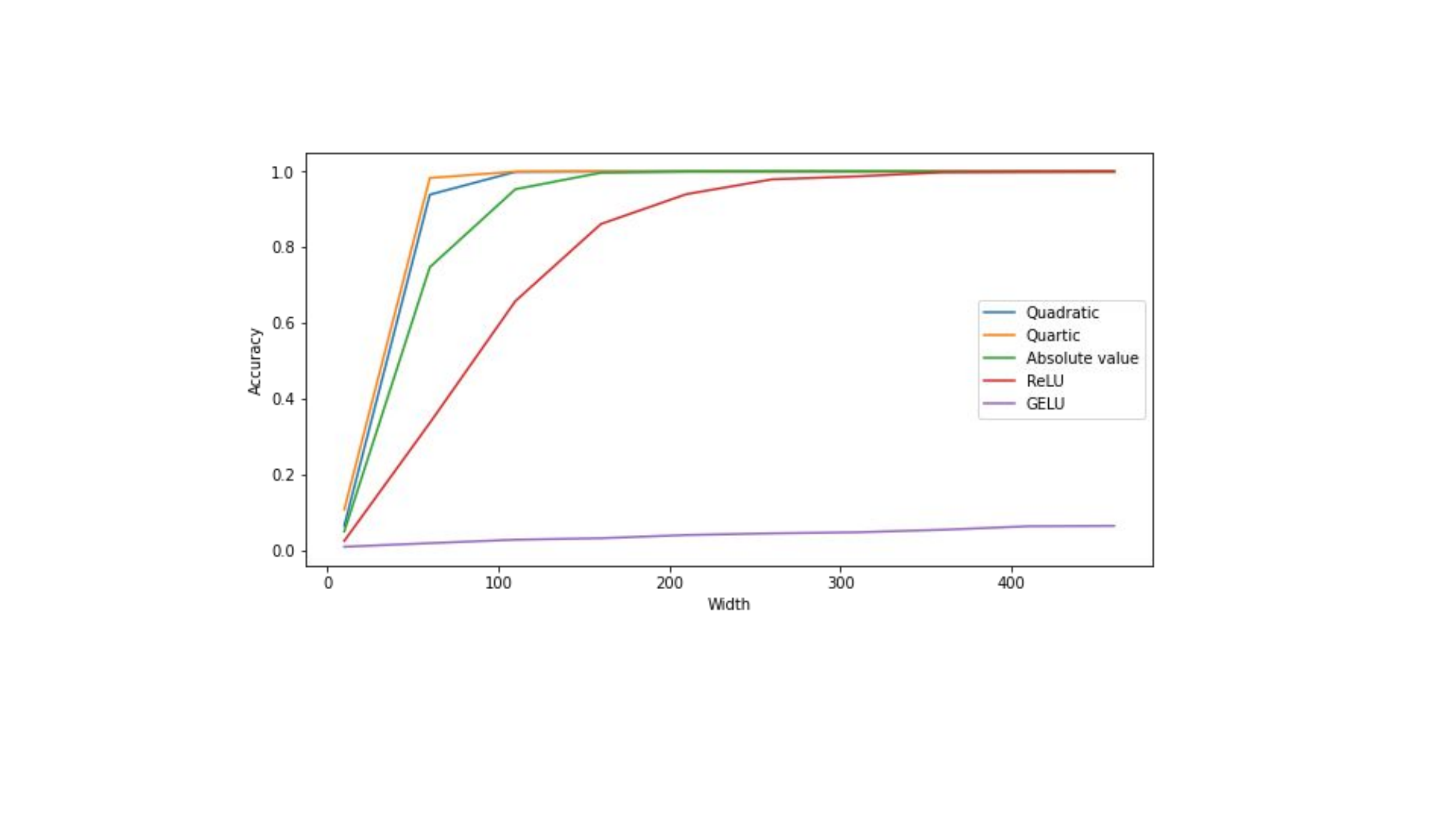}
    \caption{Accuracy for various activation functions. Test accuracy vs. width for different activation functions for $f(n,m)=n+m \,\, \textrm{mod}\,\,p$. The weights are given by \eqref{eq_sol1}-\eqref{eq_sol2}. GELU activation eventually reaches $100\%$ accuracy, but at very large width.}
    \label{fig:figA1}
\end{figure}

Remarkably, the weights \eqref{eq_sol1}-\eqref{eq_sol2} also solve the modular addition problem for networks \eqref{eq_h1}-\eqref{eq_h2} with other activation functions.  
That is, the function 
\begin{equation} \label{eq_f_relu}
    f(x) = \frac{1}{D \sqrt{N}}  W^{(2)} \phi\left(W^{(1)} x\right)\,
\end{equation}
approximates the $\delta$-function concentrated on the modular addition problem. This also holds for the generalizations discussed in the main text. We do not have an analytic proof of this fact, so we provide the evidence in Fig. \ref{fig:figA1}.

\section{Some other modular functions}

We show a few examples of the modular functions for which the exact solutions discussed in the main text apply.
\begin{itemize}
    \item $f(n,m) = n^2 + m^2 \,\, \textrm{mod}\,\,p$. Full solution is available and gives $100\%$ accuracy. The first layer weights are given by
    \begin{equation} \label{eq_sol_sq1}
    W^{(1)}_{kn} = 
    \begin{pmatrix}
    \cos \left( 2 \pi\frac{k}{p} n_1^2 +\varphi^{(1)}_k\right)
    \\
    \cos \left( 2 \pi\frac{k}{p} n_2^2 +\varphi^{(2)}_k\right)
    \end{pmatrix}\,, \qquad n=(n_1,n_2)\,,
\end{equation}
    while the second layer weights remain unmodified.
    
    \item $f(n,m) = (n+m)^2 \,\, \textrm{mod}\,\,p$. The weights in the first layer are unmodified, while the weights in the second layer are given by 
    \begin{equation}\label{eq_sol_sq3}
    W^{(2)}_{qk} = \cos \left( -2 \pi\frac{k}{p} q^{\frac{1}{2}}  -\varphi^{(3)}_k\right)\,.
\end{equation}
Note that $q^{\frac{1}{2}}$ must be understood in the modular sense, that is $r = q^{\frac{1}{2}}$ is a solution to $r^2 = q \,\, \textrm{mod}\,\,p$.
    \item $f(n,m) = nm$. We do not have an analytic solution. The activations are presented in Fig.~\ref{fig:figA4}
    
    \item $f(n,m) = n^2 + m^2 + nm \,\, \textrm{mod}\,\,p$. We do not have an analytic solution. This generalization on this function never reaches $100\%$ unless most of the data is utilized, $\alpha > 0.95$. See the learning curve in Fig.~\ref{fig:figA5}. Note that although generalization accuracy is very high: $\approx 97\%$, there is a large gap between train and test loss. This is to be contrasted with Fig.~\ref{fig:fig1}, where the gap disappears over time. 
    
    \item $f(n,m) = n^3 + nm^2 + m$. We do not have an analytic solution. The generalization never rises above $1\%$. See the learning curve in Fig.~\ref{fig:figA5}.
\end{itemize}

\begin{figure}[!h]
    \centering
    \includegraphics[width=\textwidth]{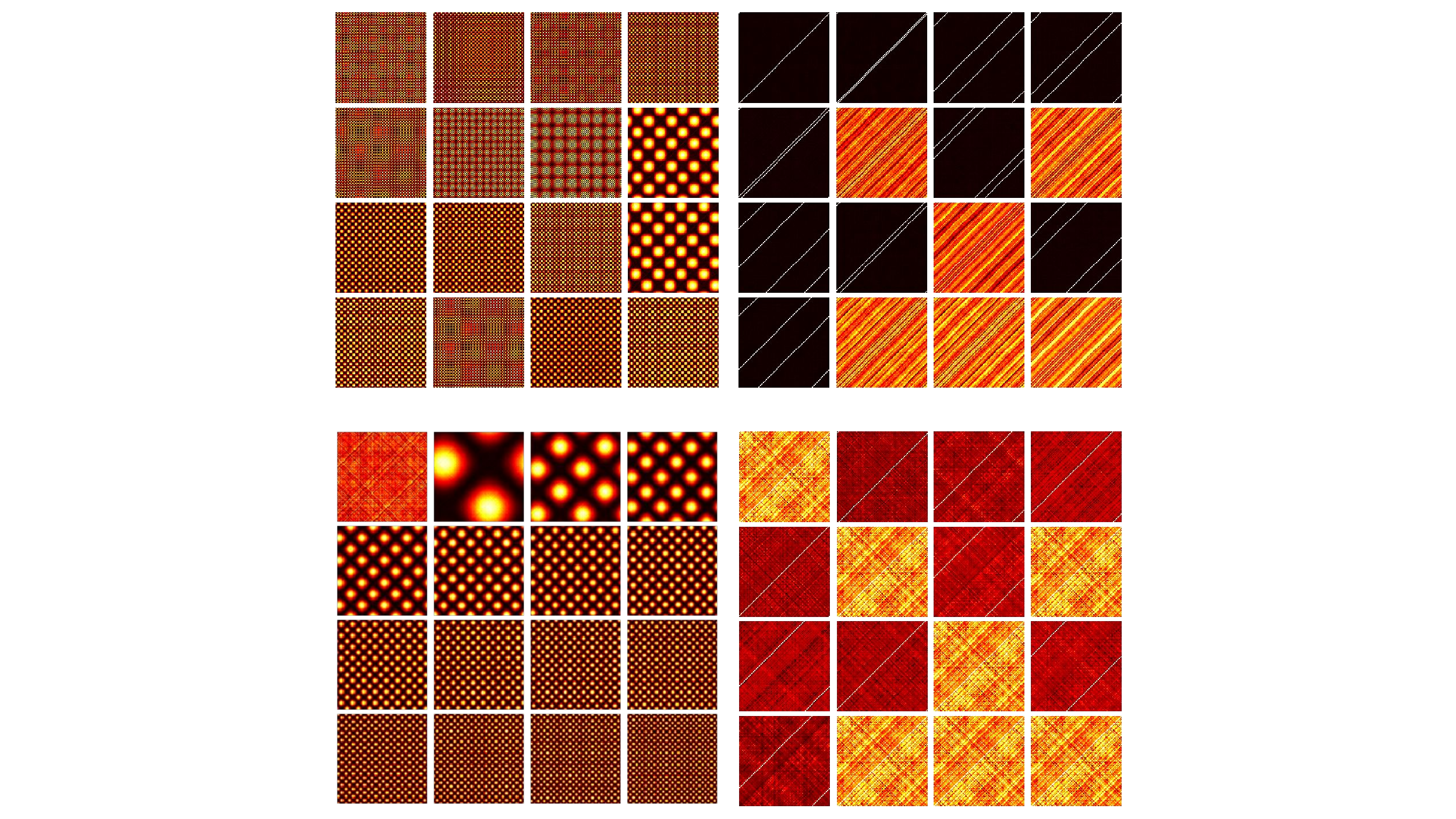}
    \caption{\textbf{Top}: Preactivations $h^{(1)}_k$ and $h^{(2)}_q$ found by the AdamW for $f(n,m) = (n+m)^2\,\, \textrm{mod}\,\,p$. Note that $h^{(1)}_k$ is the same as for $f(n,m) = (n+m)\,\, \textrm{mod}\,\,p$ as expected. \textbf{Bottom}: Analytic solution for the same function. Note that since square root is \emph{not} invertible -- because it has two branches -- the accuracy of analytic solution is $\approx51\%$. It can be clearly seen in the form of $h^{(2)}_q$: there are $4$ activation lines in the top plots and only $2$ in the bottom. Each pair corresponds to a branch of square root. The noisy preactivations $h^{(2)}_q$ correspond to the values of $q$ that cannot be represented as $(n+m)^2\,\, \textrm{mod}\,\,p$.}
    \label{fig:figA2}
\end{figure}

\begin{figure}[!h]
    \centering
    \includegraphics[width=\textwidth]{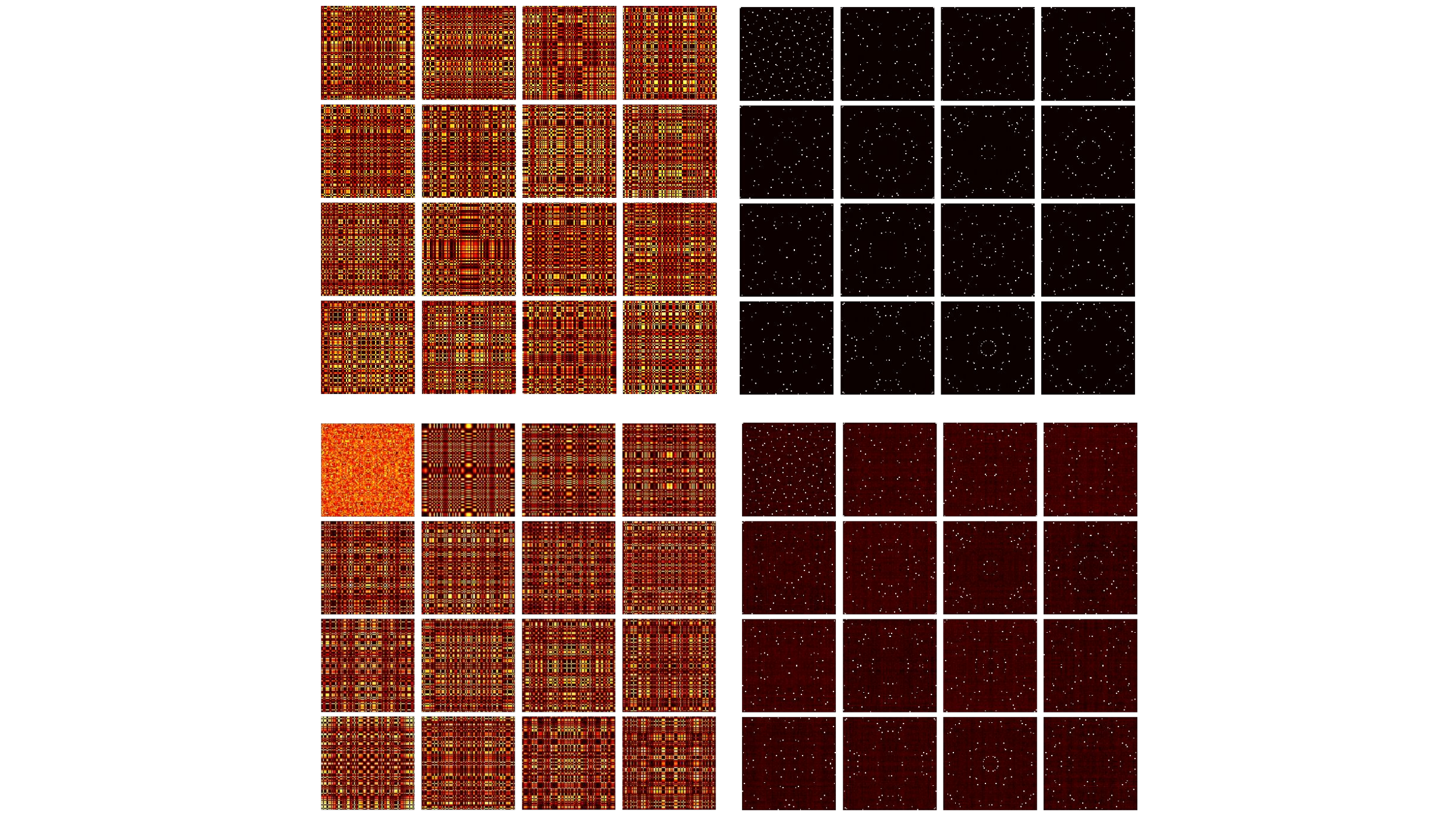}
    \caption{\textbf{Top}: Preactivations $h^{(1)}_k$ and $h^{(2)}_q$ found by the AdamW for $f(n,m) = n^2+m^2\,\, \textrm{mod}\,\,p$. \textbf{Bottom}: Analytic solution for the same function. Both solutions have $100\%$ accuracy.}
    \label{fig:figA3}
\end{figure}

\begin{figure}[!h]
    \centering
    \includegraphics[width=\textwidth]{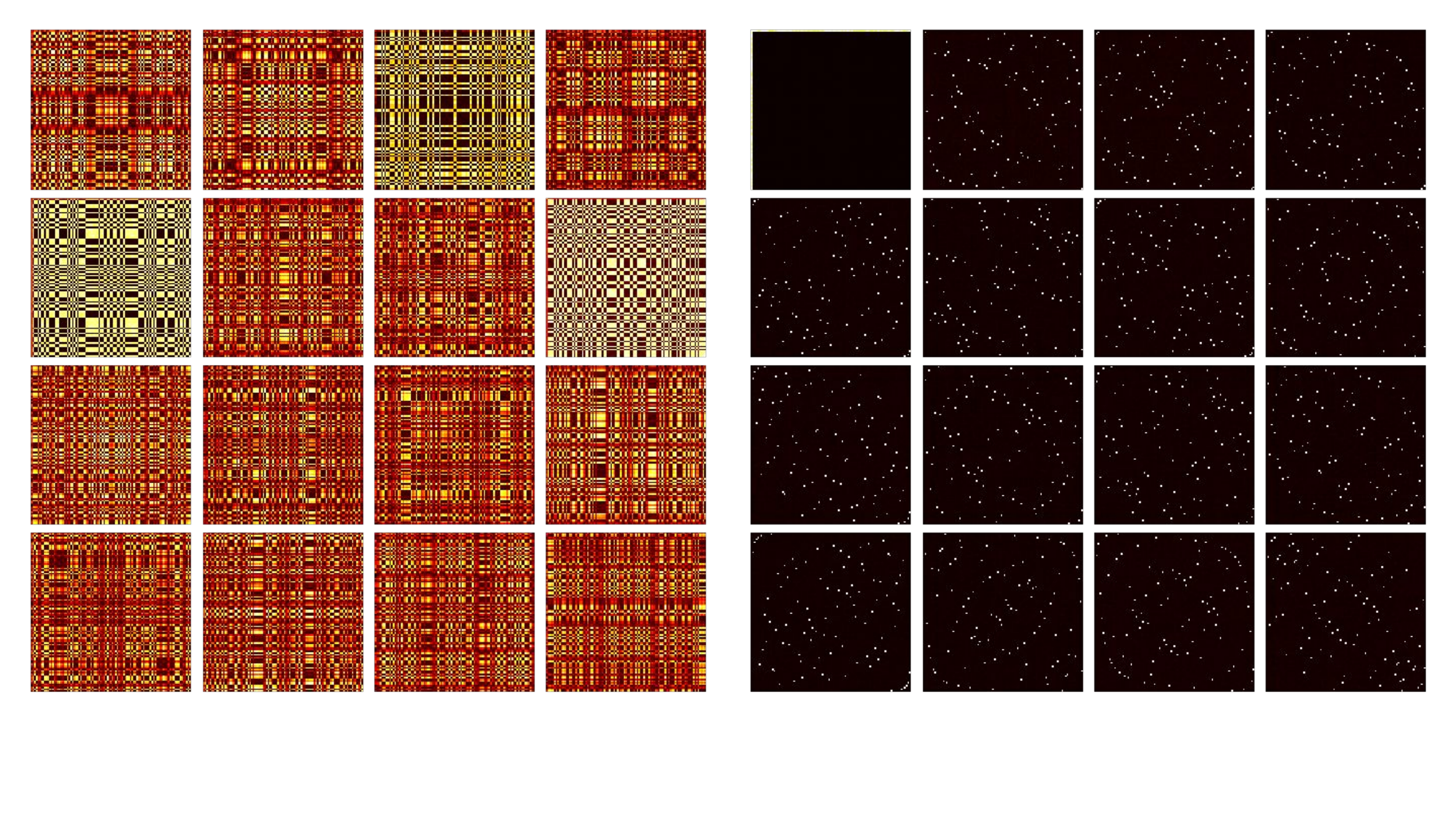}
    \caption{Preactivations $h^{(1)}_k$ and $h^{(2)}_q$ found by the AdamW for $f(n,m) = nm\,\, \textrm{mod}\,\,p$.}
    \label{fig:figA4}
\end{figure}

\begin{figure}[!h]
    \centering
    \includegraphics[width=\textwidth]{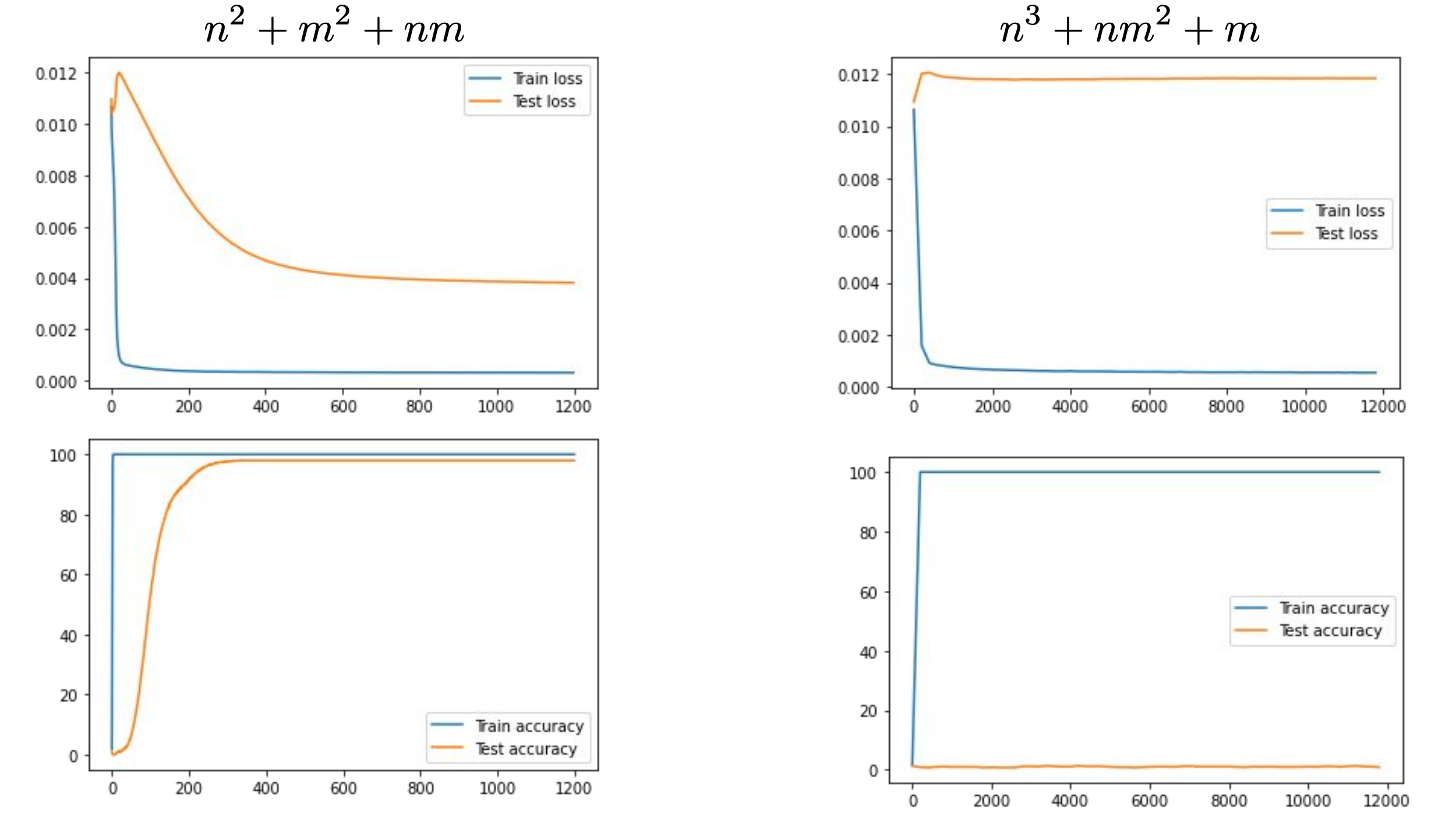}
    \caption{The learning curves for $f(n,m) = n^2 + m^2 + nm\,\, \textrm{mod}\,\,p$ and $f(n,m) = n^3 + nm^2 + m\,\, \textrm{mod}\,\,p$ at $\alpha=0.73$ and $\alpha = 0.9$ correspondingly. Note the gap between train and test loss in the former case. Although test accuracy is almost $100\%$, it is clear that the network did not grok all the right features.}
    \label{fig:figA5}
\end{figure}

We show the corresponding activations on Fig.~\ref{fig:figA2} - Fig.~\ref{fig:figA4}

\end{document}